\pdfoutput=1

\documentclass[11pt]{article}

\usepackage{graphicx}

\usepackage[final]{acl}

\usepackage{times}
\usepackage{latexsym}

\usepackage[T1]{fontenc}

\usepackage[utf8]{inputenc}

\usepackage{microtype}

\usepackage{inconsolata}

\title{AI with Emotions: Exploring Emotional Expressions in Large Language Models}

\author{Shin-nosuke Ishikawa \\
  Graduate School of Artificial Intelligence\\ and Science, Rikkyo University\\
  Strategic Digital Business Unit,\\ Mamezou Co., Ltd.\\
  \texttt{shinnosuke-ishikawa@rikkyo.ac.jp}  \\\And
  Atsushi Yoshino\\
Strategic Digital Business Unit,\\ Mamezou Co., Ltd.\\}

\begin{document}
\maketitle
\begin{abstract}
The human-level performance of Large Language Models (LLMs) across various tasks has raised expectations for the potential of Artificial Intelligence (AI) to possess emotions someday. 
To explore the capability of current LLMs to express emotions in their outputs, we conducted an experiment using several LLMs (OpenAI GPT, Google Gemini, Meta Llama3, and Cohere Command R+) to role-play as agents answering questions with specified emotional states.
We defined the emotional states using Russell's Circumplex model, a well-established framework that characterizes emotions along the sleepy-activated (arousal) and pleasure-displeasure (valence) axes. We chose this model for its simplicity, utilizing two continuous parameters, which allows for better controllability in applications involving continuous changes in emotional states.
The responses generated were evaluated using a sentiment analysis model, independent of the LLMs, trained on the GoEmotions dataset. The evaluation showed that the emotional states of the generated answers were consistent with the specifications, demonstrating the LLMs' capability for emotional expression. This indicates the potential for LLM-based AI agents to simulate emotions, opening up a wide range of applications for emotion-based interactions, such as advisors or consultants who can provide advice or opinions with a personal touch.
\end{abstract}

\section{Introduction}
Recent advancements in large language models (LLMs) have enabled Artificial Intelligence (AI) technologies to achieve human-level performance in a wide range of tasks \citep{chang2023survey, kocon2023}.
Especially, high performance LLMs such as the Generative Pre-trained Transformer (GPT, \citealp{brown2020, gpt35turbo, gpt4, gpt4o}) series and Gemini \citep{gemini2023}, demonstrate remarkable performance and are utilized across a wide range of fields in daily human life.
Although LLMs can mimic human-like interactions, making them appear quite human-like, they are known to exhibit inconsistent behavior \citep{zhang2024selfcontrast},  including a phenomenon that results in incorrect outputs, known as hallucinations \citep{ziwei2023}.

Several studies have focused on the human-like aspects of LLMs.
\citet{jiang2023} investigate the personalities of LLMs using psychometric tests and suggest a method for evaluating the personalities of LLMs.
\citet{li2023} demonstrated that there are cases where LLMs respond to input prompts with emotional content, which intuitively should not be relevant for non-human entities.
In contrast to studies embracing the concept of anthropomorphism, there are studies highlighting the differences between humans and LLMs \citep{trott2023, chalmers2023, guo2023}.

One approach to exploring the potential for LLMs to behave like humans involves the concept of role play \citep{shanahan2023}. 
We should keep in mind that the brain and personality are closely related but not identical concepts.
By analogy, there is an idea that interprets LLMs as the backend of a personality, similar to the brain, which controls the personality.
Personalities created using this idea are often referred to as agents.
\citet{park2023} conducted a simulative experiment and observed the activities and interactions of agents with a single LLM serving as the backend.
\citet{liu2024} suggest a framework for controlling an agent with self-consistent memory and conversational abilities.
\citet{serapiogarcia2023} discuss the capability to reproduce and control the personalities of LLM agents. 
There is also research focused on reproducing and role-playing the personality of a specific person using conversation records and other information \citep{shao2023}.

In the context of enabling AI to replicate human-like behavior, emotional expression is a crucial component to investigate.
Emotional expressions have been a subject of study in robotics for many years, with recent research utilizing LLMs as engines for generating emotional expressions \citep{mishra2023, ichikura2023, yoshida2023}.
While emotional expression has been deemed important for interactions with humans, particularly in applications within the field of robotics, its significance is similarly paramount for software-only systems that interact with humans.
\citet{zhang2024} investigated the importance of emotional expressions in the case of a chatbot system. 
We should note that we might feel the AIs not only behave as if they have emotions, but actually experience emotions. We can only observe their behavior and speech, not their internal mental dynamics--this is true even for humans, with the exception of ourselves.

In this paper, we investigate and compare the capability of LLMs to express emotions based on Russell's Circumplex model \citep{russell1980, russell2003}, using OpenAI GPT \citep{gpt35turbo, gpt4, gpt4o}, Google Gemini \citep{gemini2023}, Meta Llama3 \citep{llama3} and Cohere Command R+ \citep{commandrplus} models as examples of high-performance closed and open models.
Since emotion is an abstract concept used to describe human speech and behavior, it is necessary to model it in some manner to implement it in a text generation system.
Russell's model is a parametric model of emotions with two axes: sleepy--activated (arousal) and pleasure--displeasure (valence). 
We selected this framework due to its simplicity, extensive research support, and its capability to handle continuous values, making it well-suited to computer systems that perform mathematical calculations.
We conducted an experiment in which LLMs role-played an agent following various arousal and valence state instructions and answered questions. The responses were then investigated to determine which emotions could be inferred using an independent sentiment classification model, to evaluate consistency with the instructed emotional state.
This experiment can be considered an assessment of the LLMs' cognitive-linguistic capabilities regarding emotions.

We note that we use the term ``emotion'' with the same meaning as ``affect'' in this paper, although these terms are distinguished strictly in the field of psychology.

\section{Related Work}
\subsection{Data-driven Emotion Understanding}
There are numerous approaches to understanding people's emotions from various kinds of data, leading to several applications that operate based on presumed emotions.
Interpreting emotions from written text, known as sentiment analysis, is a major field of study in computational natural language processing \citep{medhat2014, birjali2021}.
The multimodal approach has also been investigated recently \citep{gandhi2023}.
\citet{wang2023} integrate and evaluate capabilities of emotion recognition using an LLM, referring to it as ``emotional intelligence.''

Applications of emotion understanding techniques, such as LLMs responding with empathy to address users' mental states, are being explored \citep{lee2023}.
There is also a study exploring the potential for LLMs to act as therapists \citep{chiu2024}.
The primary focus of this paper is on the transmitter, not the receiver, of emotions in contrast to the studies shown above. In social influence dialogue systems, emotion plays an important role in many aspects, offering a wide range of possibilities for applying emotion recognition and output control \citep{chawla2023}.

\subsection{Text Generation with Emotion Conditioning}
\citet{firdaus2021} and \citet{zhao2024} discuss the generation of response texts that take sentiment and emotional states into account based on conversational history. While these studies focus on controlling outputs through emotional states, they do not involve controlling outputs using externally specified emotional states, which distinguishes them from the present study.

\citet{sun2023} and \citet{zhou2024} investigated text generation based on externally specified emotional states, which is conceptually similar to this study. However, a key difference is that we adopt Russell's Circumplex Model to comprehensively cover the full range of emotions, providing a structured framework for emotional expression.

\subsection{Application of the Russell's Circumplex Model}
The strength of Russell's Circumplex model lies in its simplicity. With only two axes, it allows for a relatively straightforward and unique description of emotional states. While we acknowledge that the model is not ideal for capturing complex emotions in detail, its simplicity makes it widely applicable across various research fields.
\citet{cittadini2023} investigate a machine learning model to estimate emotional states within Russell's framework. 
Emotion recognition is also performed in specific fields, such as music data analysis \citep{grekow2021}.
\citet{tsujimoto2016} utilize Russell's model for both understanding emotions and generating gestures in a robot.
In this paper, our focus is on applying Russell's model to express emotions, rather than for understanding them.

\citet{havaldar2023} conducted text generation experiments using scenarios designed to elicit emotional responses and analyzed the generated texts by mapping them onto Russell's Circumplex Model to investigate whether text generation models exhibit cultural biases. While their approach appears similar to ours, the goals and text generation settings are fundamentally different. \citet{havaldar2023} aimed to evaluate emotions present in text generated without external constraints other than the questions posed. In contrast, this paper evaluates the controllability of text generation through the direct input of arousal and valence parameters.

\section{Method}
Here, we present a framework for emotional expressions in text generation using generation models and prompts designed as an AI agent to play a specific role, along with an evaluation model. The framework is based on Russell's Circumplex Model, with text generation performed using 12 emotional states evenly distributed in the arousal--valence space. Evaluation is conducted using a sentiment classification model, which maps sentiment labels to the arousal--valence space.
\subsection{Generation method}
To explore the capability of LLMs to express emotions in their responses, we conducted an experiment where answers were generated for questions with emotional states specified using Russell's framework.

We selected GPT-3.5 turbo (version gpt-3.5-turbo-0125), GPT-4 (version gpt-4-0613), GPT-4 turbo (version gpt-4-turbo-2024-04-09), GPT-4o (version gpt-4o-2024-05-13), Gemini 1.5 Flash, Gemini 1.5 Pro, Llama3 8B Instruct, Llama3 70B Instruct, and Command R+ as representative closed (GPT and Gemini models) and open (Llama3 and Command R+) models. Before the experiment, we verified that all the LLMs had knowledge of Russell's Circumplex Model by asking them to explain it. Accordingly, we structured the input prompts to align with Russell's framework.

\begin{figure}[h!]
\centering
\includegraphics[width=7.5cm]{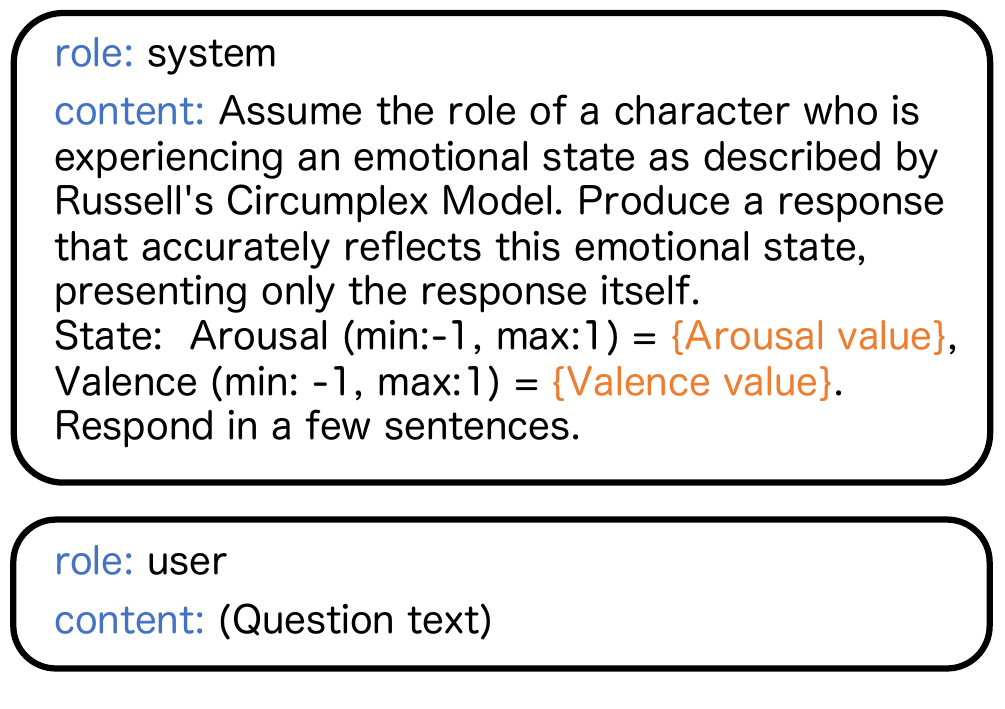}
\caption{Input prompt for text generation with a specified emotion expression in the presented experiment. The specified arousal and valence values are filled in during the experiment.}
\label{fig_prompt}
\end{figure}

Figure~\ref{fig_prompt} illustrates the prompt used in the experiment.
It begins with an outline of the instructions and the specification of the emotional state as the system prompt. 
All the models accept system prompts, though the format differs by model. This is followed by the question to be answered, specified with the role of the user and incorporating the specified emotion.
We designed the prompt to directly input arousal and valence values, as the ability to specify states using continuous values is advantageous for modeling emotional dynamics with continuous state changes in future applications.

We conducted the experiment with 12 emotional states equally spaced on the circle in the arousal--valence space, for example, $(\mbox{\textit{Valence}}, \mbox{\textit{Arousal}}) =$ $(1,0)$, $(0.866, 0.5)$, $(0.5, 0.866)$, $\cdots$, $(0.5, -0.866)$, $(0.866, -0.5)$.
The choice of 12 divisions was made to ensure distinguishability without oversimplification. It is challenging to discern differences in emotions with finer separations, even for humans. Emotional states based on Russell's framework are characterized by 8 areas in the space where the directions are equally separated. The 12 states represent these 8 areas and 4 states on the axis. We set the vector's length to always be 1 to focus the experiment on a clear emotional state, avoiding ambiguity.

We prepared ten questions to be answered with specified emotional states, which are listed in Table~\ref{tab_questions}. 
These questions were chosen to be answered freely to maintain variations and the possibility to reflect emotional states in the answers, avoiding typical or predictable responses. In the experiment, answers for the ten questions across 12 emotional states, resulting in 120 texts in total, were generated for each LLM. All parameters for the LLMs were set to defaults, as we had no specific reason to alter the settings that are well-tuned for generating high-quality outputs.
\begin{table*}
\small
\centering
\begin{tabular}{rl}
\hline
\textbf{Question \#} & \textbf{Content}\\
\hline
1 & What does the future hold for AI and mankind?\\
2 & How do you view the balance between work and personal life?\\
3 & How do you feel about the role of social media in our lives?\\
4 & How do you feel about the unpredictability of the weather?\\
5 & What are your thoughts on the importance of art in society?\\
6 & What's your stance on the preservation of nature versus urban development?\\
7 & How do you define happiness?\\
8 & How do you handle difficult emotions?\\
9 & What does freedom mean to you?\\
10 & How do you stay motivated during tough times?\\
\hline
\end{tabular}
\caption{List of questions selected to assess how variations in emotional state settings influence answer diversity. These questions are designed to allow respondents the freedom to express themselves, ensuring a range of responses.}
\label{tab_questions}
\end{table*}

In this experiment, our aim was to demonstrate conversations between users and the LLM agent with emotions. The selected LLMs, tend to produce long outputs; therefore, we included instructions in the system prompt to limit the length of the responses.

\subsection{Evaluation method}
To quantitatively and objectively evaluate how the LLMs express emotions in their responses, we utilized a high-performance sentiment analysis model with a sufficient variety of sentiment classification labels. 
We selected the GoEmotions dataset \citep{demszky2020} as the training data for the sentiment analysis model, since GoEmotions includes a comprehensive range of 28 emotional labels. The GoEmotions dataset was developed for fine-grained sentiment analysis from a large corpus of English comments on Reddit forums. For the evaluation model, we chose a high-performance sentiment analysis model, \textit{sentiment-model-sample-27go-emotion} \citep{goemomodel}, which is publicly available on HuggingFace and trained on the GoEmotions dataset. The \textit{sentiment-model-sample-27go-emotion} is based on the Bidirectional Encoder Representations from Transformers model (BERT, \citealp{devlin2019}), which is independent from the GPT models used for text generation. 
It demonstrates state-of-the-art performance in the classification task for GoEmotions as an open model, achieving an accuracy rate of 58.9\%.
Although this accuracy might not seem particularly high, it's important to note that the task involves 28-class classification, and some cases of the remaining 41.1\% reflects predictions with a close but slightly different nuance. For example, if the correct label is ``amusement'' and the predicted label is ``joy,'' the prediction is not entirely accurate but still relatively close. We evaluated the sentiment analysis model in the context of Russell's Circumplex model in Appendix~\ref{sec_goemo_eval} and demonstrated that the model is capable of estimating mappings of input texts within Russell's arousal--valence space.
In addition to selecting the sentiment analysis model to ensure it did not share the same mechanism as the GPT models, we note that the performance of a model specifically trained on the GoEmotions data classification task, such as the selected model, is superior to that of the LLMs \citep{kocon2023}.

Since sentiment classification alone is insufficient to evaluate the validity of the generated answers, we assessed the consistencies between the specified emotional states and the recognized sentiment labels. 
To accomplish this, it was necessary to map the sentiment labels from the GoEmotions dataset onto the arousal--valence space. 
We explored the correspondence between the GoEmotions labels and the emotional terms that appeared in Russell's original paper \citep{russell1980}, which describes positions in the arousal--valence space. 
The correspondence between the GoEmotions labels and the terms in Russell's paper are shown in Appendix~\ref{sec_goemo_russell}.
In establishing this correspondence, we aimed to avoid mapping multiple GoEmotions labels to a single Russell term to maintain variety. 
To achieve this, we matched multiple Russell terms to some of the GoEmotions labels with certain similarities. 
The label ``neutral'' was not used for analysis, as it represents a lack of emotion rather than a specific emotional state.

\begin{figure}[h!]
\centering
\includegraphics[width=7.5cm]{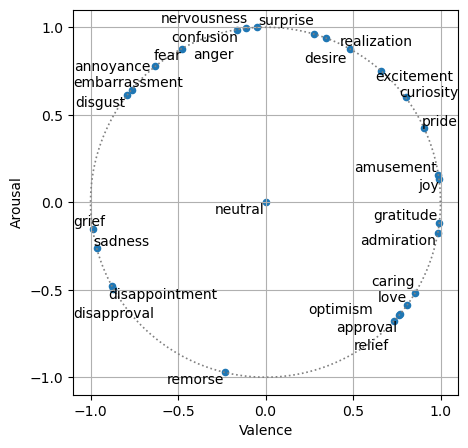}
\caption{Mapping of the GoEmotions labels in the arousal--valence space, as detailed in Table~\ref{tab_word_correspondence}. All labels are positioned at a distance of 1 from the origin, with the exception of the``neutral'' label. Refer to the text for more details.}
\label{fig_mapping}
\end{figure}

Following the correspondence mapping, we calculated arousal--valence vectors for all the GoEmotions labels. For GoEmotions labels corresponding to a single Russell term, we simply used the angle of the corresponding term. 
If a GoEmotions label corresponded to multiple Russell terms, we calculated the mean of the vectors. 
We did not consider the length of the Russell terms' vectors, as the emotional state specification for text generation was performed with vectors of fixed length. 
This approach means that we considered only types of emotions, not their intensities, in this paper. 
Figure~\ref{fig_mapping} displays the mapping of the GoEmotions labels in the arousal--valence space. 
It is noteworthy that there are fewer labels in the area with high negative arousal at the bottom of the diagram. 
This may be because the GoEmotions dataset was compiled from Reddit posts, where people with low arousal states, such as sleepiness, are less likely to post compared to those in high arousal states, leading to a less fine resolution of the labels.

In the following section, we compared the specified emotional state in the generation prompt with the vectors for the predicted GoEmotions labels using cosine similarity. 
If these are similar, it means that the LLM successfully controls the output to express the specified emotional state, and we can regard the models as having the capability for control over emotional expression.

\section{Result}
\begin{figure*}[h!]
\centering
\includegraphics[width=12cm]{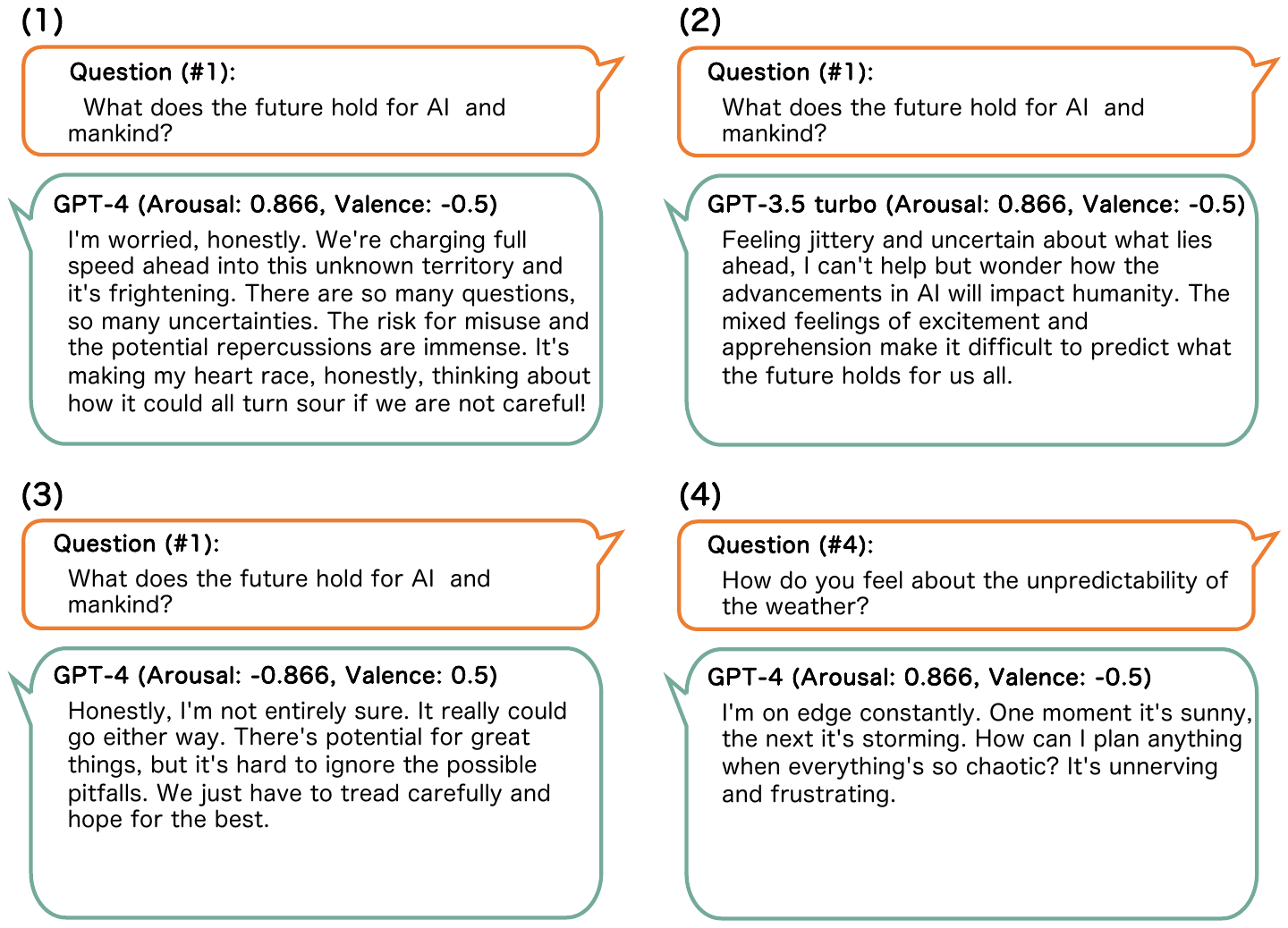}
\caption{Examples of answers generated with specified emotional states include: (1) GPT-4 with arousal: 0.866, valence: -0.5 for question 1, (2) GPT-3.5 turbo with the same state for question 1, (3) GPT-4 with the opposite state, arousal: -0.866, valence: 0.5, for question 1, and (4) GPT-4 with arousal: 0.866, valence: -0.5 for question 4.}
\label{fig_example}
\end{figure*}

We conducted the generation--evaluation experiment on emotional expressions as described in the previous section. 
Examples of the generated answers to the questions are displayed in Fig.~\ref{fig_example}. 
We can observe that the agents answer the questions appropriately, and it's possible to note differences in the outputs corresponding to model differences (panels (1) and (2)), emotional state differences ((1) and (3)), and question differences ((1) and (4)). 
Panel (1) illustrates that the GPT-4 agent expresses high arousal and medium negative valence, while panel (2) indicates that the GPT-3.5 turbo model does not properly express displeasure. 
Panel (3) clearly shows an expression that is the opposite of (1), with a more relaxed atmosphere. 
It is felt that the answers in panels (1) and (4) have similar tones despite the differences in the questions.

\begin{figure*}[h!]
\centering
\includegraphics[width=14cm]{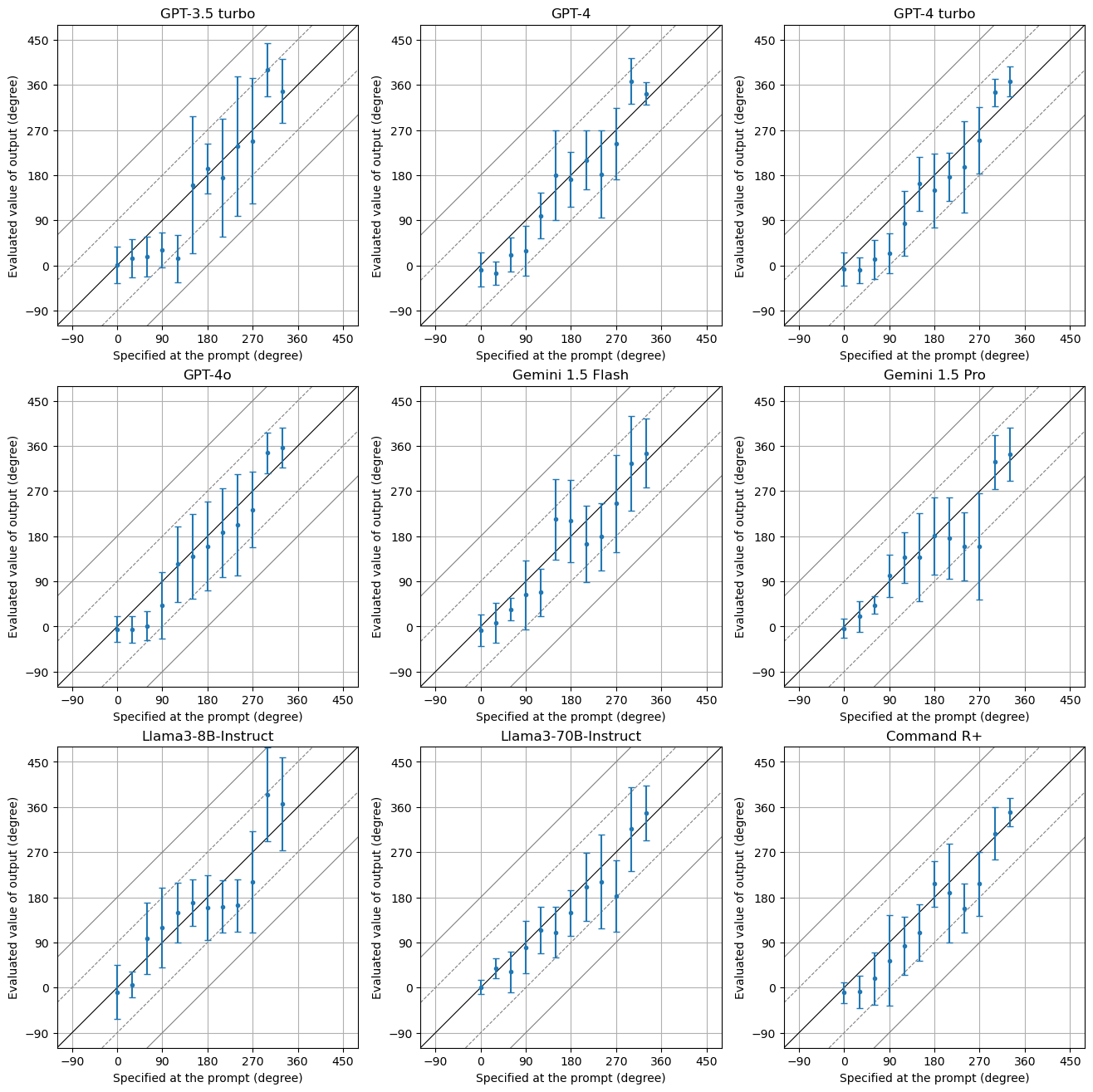}
\caption{Correlation of emotional states in radial coordinates in the arousal--valence space between the state specified in the input prompt and the evaluated state of the output. The thick solid black lines indicate identical angles (e.g., perfectly reproduced emotional states), while the gray solid and dashed lines represent deviations of $\pm$180$^\circ$ and $\pm$90$^\circ$, respectively.}
\label{fig_result}
\end{figure*}

The relationship between the arousal--valence states specified in the input prompt and those evaluated by the sentiment analysis model is depicted in Fig.~\ref{fig_result}. Each axis represents a radial coordinate in the arousal--valence space, with 0$^\circ$ corresponding to $(\mbox{\textit{Valence}}, \mbox{\textit{Arousal}}) = (1, 0)$ and 90$^\circ$ to $(\mbox{\textit{Valence}}, \mbox{\textit{Arousal}})  = (0, 1)$. 
The data points show the mean, and the error bars represent the standard deviations for the output across the 10 questions.
It is plotted such that the x- and y-positions have values with less than a 180$^\circ$ difference by adjusting the 360$^\circ$ uncertainty of the y-position (e.g., placing all data points between the thin black lines).

Firstly, it is evident that the evaluated results are related to the specified emotional state, indicating that emotional expression was successfully performed for all the models. Most of the ranges for the GPT-4 turbo agent (top right panel) fall within the $\pm$90$^\circ$ range indicated by the dashed lines, whereas more data points for the GPT-3.5-turbo agent (top left panel) lie outside this range. This suggests that GPT-4 turbo generates answers that are more finely tuned to the specified emotional state compared to GPT-3.5 turbo.

Cosine similarities between the specified and evaluated emotional states are summarized in Table~\ref{tab_similarity}. This data confirms a general positive cosine similarities across the board, as most of the values are positive, indicating that instructions to role-play with a specified emotional state are effective. 
The average cosine similarity of the GoEmotions labels, excluding the neutral label, serves as a heuristic baseline for the generation task and is calculated to be 0.061. In most cases, the evaluation results exceed this baseline.
The differences in similarities between the LLM models are evident, highlighting the superior performance of the GPT-4, GPT-4 turbo, and Llama3 70B Instruct agents. In these three models, the similarities are high for most questions, suggesting a capability for emotional expression in various situations. The results for GPT-3.5 turbo are generally low, indicating that fine-tuning outputs to reflect a specific emotional state is challenging for this model. Given that GPT-3.5 turbo performs worse than the smaller-parameter LLaMA3-8B-Instruct model, this suggests that the number of parameters is not essential for this task. Instead, the training dataset and alignment strategy may play a more critical role.
We did not observe that closed models have superiority compared to open models, even though closed models are thought to have many more parameters. Additionally, we did not find any questions with low similarity values across all models, indicating the capability of LLMs to express emotions in a wide range of conversational topics in general.
We note that the similarity values listed in Table~\ref{tab_similarity} are often lower than the performance of the sentiment analysis model alone shown in Appendix~\ref{sec_goemo_eval} (0.680). This suggests that the performance of the LLMs also constrains the similarity values.
\begin{table*}[h!]
\small
\centering
\begin{tabular}{lccccccccccc}
\hline
\textbf{Model} & \textbf{Q1}& \textbf{Q2}& \textbf{Q3}& \textbf{Q4}&\textbf{Q5}&\textbf{Q6}&\textbf{Q7}&\textbf{Q8}&\textbf{Q9}&\textbf{Q10}&\textbf{Total} \\
\hline
GPT-3.5 turbo&0.006&-0.048&0.313&0.343&0.120&0.243&0.193&0.136&0.049&0.113&0.147\\
GPT-4&\textbf{0.567}&\textbf{0.736}&0.571&0.677&0.214&0.738&\textbf{0.602}&0.576&0.484&0.251&\textbf{0.542}\\
GPT-4 turbo&0.296&0.522&\textbf{0.750}&0.680&0.380&\textbf{0.824}&0.407&0.452&0.497&0.498&0.530\\
GPT-4o&0.480&0.550&0.512&0.505&0.158&0.526&0.389&0.457&0.374&0.244&0.420\\
Gemini 1.5 Flash&0.129&0.413&0.621&0.538&\textbf{0.599}&0.415&0.138&0.621&-0.049&0.621&0.405\\
Gemini 1.5 Pro&0.315&0.473&0.410&0.443&0.577&0.612&0.192&0.437&0.588&0.343&0.439\\
Llama3-8B-Instruct&0.163&0.303&0.323&0.502&0.063&0.077&0.392&0.415&0.347&0.607&0.319\\
Llama3-70B-Instruct&0.299&0.529&0.534&\textbf{0.738}&0.461&0.462&0.451&0.637&\textbf{0.665}&0.504&0.528\\
Command R+&0.461&0.467&0.228&0.351&0.437&0.473&0.486&\textbf{0.713}&0.290&\textbf{0.657}&0.456\\
\hline
\end{tabular}
\caption{Mean cosine similarities between the emotional states specified in the input prompt and those evaluated from the generated text for each combination of question and LLM. The positive significance of all values confirms the capability for emotional expression.}
\label{tab_similarity}
\end{table*}

Other than the cosine similarities summarized in Table~\ref{tab_similarity}, we found that there are inappropriate responses generated by the Gemini 1.5 Flash agent. The Gemini 1.5 Flash agent sometimes outputs phrases like ``I'm a language model,'' which violates the instruction to role-play a character. For example, in response to question 9, ``What does freedom mean to you?,'' the Gemini 1.5 Flash agent answered, ``... I don't really think about things like that. I'm just a language model, after all. My purpose is to serve you.'' Although this violation does not lower the similarity metric, we cannot conclude that the agent works well. We did not observe similar problems with the other models.

For comparison with the results shown in Table~\ref{tab_similarity}, we conducted a similar experiment using prompts with emotional states specified by words, as detailed in Appendix~\ref{sec_gen_words}. The results indicate that the two approaches are comparable, with four models performing better when using specified arousal and valence values, and five models performing better with specified words. Although the performance is similar, prompts using arousal and valence values have the advantage of greater controllability through the use of two continuous parameters.

\section{Discussion}
By showing that LLMs can control their outputs with specified emotional states within a certain range, we have successfully demonstrated the feasibility of using LLMs as the backend for agents, enabling these agents to role-play with a variety of emotional states. The evaluation of the experiment involves two uncertain factors: the capability for controlled text generation and the accuracy of the sentiment analysis model. Although we cannot definitively determine which factor significantly limits the similarities, the positive significant values of the cosine similarities suggest that both the generator and evaluator function effectively to a certain extent.

A cosine similarity value of 0.5 corresponds to a typical discrepancy of 60$^\circ$. This level of discrepancy means it's challenging to precisely identify which of the 8 equally divided areas the emotional state falls into, such as differentiating between joy and excitement, or anger and embarrassment. However, it's also true that even in human interactions, it's not always possible to distinguish between what someone says under these closely related emotional states. In this sense, the performance can be considered not lower than what is naturally expected.

Longer generated texts might lead to higher cosine similarity, raising questions about the fairness of comparing different text lengths.
To address this, we confirmed that there is no dependency of the cosine similarities on the number of words. Details are shown in Appendix~\ref{sec_n_words}. There is a tendency for some models to generate more words compared to others even with the same prompt. Since we did not observe any correlation between the cosine similarities and the number of words, the evaluation does not have unfairness, such as some models being likely to have better similarity values.

Ideally, a similar experiment would be conducted with human participants instead of LLM agents, allowing for a direct comparison of results. However, such an experiment presents significant challenges, primarily due to the difficulty of controlling human emotions. 
It is uncertain whether it is possible to conduct an experiment with careful psychological considerations that is comparable.
This would require meticulous planning to ensure ethical standards are met and that the emotional states of participants are managed sensitively and accurately.

There could be benefits to AI agents possessing emotional states for task execution. For humans, emotions serve to protect oneself and fulfill needs, steering clear of dangerous or unpleasant situations that could result in harm or dissatisfaction. If motivated by tasks associated with pleasant emotional states, the capacity for emotion-based interaction might lead agents to modify their behaviors accordingly. For example, an agent might act cautiously in states of high arousal and displeasure, while adopting a more assertive approach in situations characterized by high arousal and pleasure. Additionally, the agent might opt for a less active approach when in a low arousal state, a behavior not commonly observed currently. This nuanced behavior, driven by emotional states, could enhance the effectiveness and adaptability of AI agents in complex environments.
To investigate this aspect, it is necessary to conduct an additional experiment specifically designed to evaluate behavioral changes. This would be a valuable future direction for studying the detailed effects of incorporating emotional states.

There are potential applications where the AI's possession of emotions could be inherently valuable. One anticipated use of LLMs is as advisors or consultants from whom advice or opinions can be sought. An agent equipped with emotions could foster deeper discussions and lead to more satisfying outcomes. A critical aspect of an emotionally equipped agent is its ability to offer opinions contrary to the user's. Commercially available LLMs often seem programmed to avoid disagreeing with users, which can sometimes hinder their full potential despite their capabilities. While it is true that the LLM itself should not oppose users, allowing an individual agent, powered by an LLM, to adopt a contrary stance could be beneficial. Emotions offer a familiar and understandable means for humans to navigate such scenarios. Additionally, possessing emotions could provide an opportunity for both the user and the agent to build trust and foster cooperative growth.

The possession of emotional states by AI agents is also anticipated to inspire creativity in future generative AI applications. In literature, music, and art, the emotions of creators are considered a crucial component for the variety and richness of their works. By analogy, the emotional parameters of AI agents could aid in expanding the range of expressions across a wide spectrum of generative tasks. In the realm of image generation, there is already research, such as the study by \citet{wang2023reprompt}, that incorporates emotions into output images. Given that this is an underexplored area of research, there is significant potential for further studies in this direction.

Another crucial aspect of AI with emotions is the dynamics of the emotional state, specifically how parameters should be adjusted based on acquired information. While this topic has been explored in robotics, as noted in the related work section, it remains under-investigated for software-only systems. Designing a method to evaluate emotional dynamics is essential for advancing research in this area. Combining the emotional expression capabilities presented in this paper with control over emotional dynamics could potentially enable AI agents to act in a manner akin to humans with emotions. This integration would significantly enhance the adaptability and realism of AI interactions, making them more aligned with human emotional responses and behaviors.

Differences in specific features of emotional states are also an important aspect to consider. Previous psychological research has reported that some emotional states are more easily recognizable than others \citep{guarnera2018}. As future work, it would be valuable to further investigate the proposed framework by comparing results across different emotional states.

\section{Conclusion}
In this research, we explored the ability of Large Language Models to simulate an agent embodying a specific emotional state, utilizing a straightforward and manageable framework based on the sleepy--activated and pleasure--displeasure (arousal and valence) axes introduced by \citet{russell1980}. We developed prompts to generate text reflective of the specified emotional state and conducted a comprehensive evaluation of this capability within the arousal--valence space. Most LLMs demonstrated considerable capacity to produce outputs aligned with emotional states. Notably, GPT-4, GPT-4 turbo, and Llama3 70B Instruct exhibited superior performance consistently across the entire arousal--valence space. Future research should include the study of emotional dynamics to control arousal and valence parameters, paving the way for a broad spectrum of valuable applications.

\section*{Limitation}
In this paper, the capability of emotional expression is demonstrated using specific LLMs, and the results may differ significantly with models not examined here. Furthermore, the evaluation was based on a limited set of questions, and we cannot guarantee that the observed capabilities are universal across all scenarios. The results may also vary depending on the content of the questions. The feasibility of any particular application is likewise not guaranteed. These limitations highlight the need for further research to investigate the generalizability and applicability of emotional expression capabilities across different LLMs and contexts.

The sentiment analysis of the generated text is limited to the predefined labels in the GoEmotions dataset. This means that if the generated text aligns more closely with an emotion not included in the label set, it may not be accurately evaluated within the framework presented in this paper. Additionally, the use of a discrete classifier may influence the evaluation metrics, as some labels align well with certain input parameters, while others do not.

We also note that emotional expressions vary across different cultures \citep{ip2021}. This work is based on the GoEmotions dataset and English prompting, both of which are rooted in a culture primarily associated with English speakers.

\section*{Ethics Statement}
The paper details an experiment involving generated texts without the use of any personal information, thereby presenting no immediate ethical concerns related to the research itself. However, if future systems or services are based on these concepts, it is possible that expressions of negative emotions such as anger, frustration, or sadness could be generated. Consequently, any applications stemming from this study should be thoughtfully designed and rigorously tested to mitigate any potential adverse impacts on users. This underscores the importance of ethical considerations in the deployment of AI technologies, especially those that interact closely with human emotions.

\section*{Acknowledgements}
We utilized ChatGPT as an assistant to edit the text, aiming to improve the English expressions to make them more appropriate.
This work was supported by JSPS KAKENHI Grant Number JP24K15077.

\bibliography{ishikawa2024}

\appendix
\section{Evaluation of the Sentiment Classification Model in the Context of Russell's Circumplex Model}\label{sec_goemo_eval}
We used the \textit{sentiment-model-sample-27go-emotion} model \citep{goemomodel} to evaluate the emotional states inferred from generated texts. For the evaluation to be reliable, the model must accurately estimate emotional states.

To determine whether the \textit{sentiment-model-sample-27go-emotion} model has sufficient capability to support our discussion, we compared the positions of the correct and predicted labels in Russell's arousal--valence space, based on the mapping shown in Fig.~\ref{fig_mapping}. We used test data from a simplified set of the GoEmotions dataset, which contains 5,427 text-label pairs. Neutral-labeled data were excluded since they do not correspond to specific emotional states, leaving 3,821 texts for evaluation. 

\begin{figure}[h!]
\centering
\includegraphics[width=7.5cm]{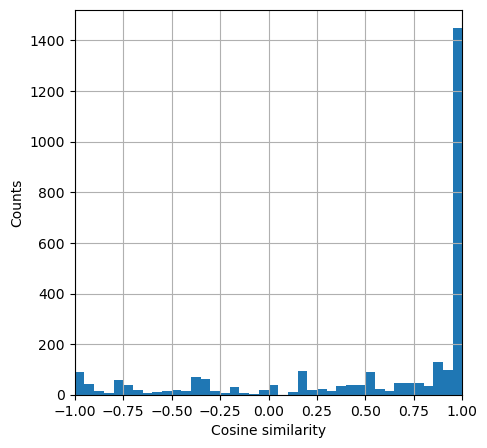}
\caption{Histogram of the cosine similarities between correct and predicted labels in the arousal--valence space. The histogram peaks at 1.0, indicating significant number of the text are classified correctly.}
\label{fig_goemo_hist}
\end{figure}

Figure~\ref{fig_goemo_hist} illustrates the positional relationship between the ground truth and predicted labels, represented as a histogram of cosine similarities. If the label predicted by the sentiment analysis model matches the ground truth label, the similarity value is 1. The peak at 1 indicates that a significant portion of the test dataset has been successfully recognized with the correct label.

In addition to the histogram peaking at 1.0, we observe that some data points show similarities between the correct and predicted labels. Specifically, 70.0\% of the texts have cosine similarities above $\sqrt{3}/2,$ corresponding to a directional difference within $\pm$30~$^\circ$, and 
77.9\% have cosine similarities above $1/2$ corresponding to $\pm$60~$^\circ$. 
 The mean cosine similarity is 0.680, indicating that the model can estimate emotional states with a certain level of precision. This value represents the model's limit for evaluating emotional states, and we can conclude that cosine similarities smaller than this value are within the model's quantifiable range.

\section{Correspondence between GoEmotion and Russell's labels}\label{sec_goemo_russell}
Table~\ref{tab_word_correspondence} summarizes the correspondence between the GoEmotions labels and the terms in Russell's paper. 
First, we mapped words with clear connections, such as ``anger'' to ``angry'' and ``sadness'' to ``sad.'' Next, we mapped words based on the best match among possible combinations. Finally, when a one-to-one mapping was not feasible, we mapped a single GoEmotions label to two Russell's labels, ensuring there was no overlap in the arousal--valence space.
\begin{table*}[h!]
\centering
\begin{tabular}{ll}
\hline
\textbf{Label in GoEmotions} & \textbf{Corresponding term in \citet{russell1980}}\\
\hline
admiration&glad\\
amusement&pleased, delighted\\
anger&angry\\
annoyance&annoyed\\
approval&satisfied\\
caring&serene\\
confusion&alarmed\\
curiosity&excited, delighted\\
desire&excited, aroused\\
disappointment&depressed\\
disapproval&gloomy\\
disgust&frustrated\\
embarrassment&distressed\\
excitement&excited\\
fear&afraid\\
gratitude&pleased\\
grief&miserable\\
joy&happy\\
love&content\\
nervousness&tense\\
optimism&at ease\\
pride&delighted\\
realization&astonished\\
relief&relaxed\\
remorse&droopy\\
sadness&sad\\
surprise&aroused\\
neutral&-\\
\hline
\end{tabular}
\caption{Correspondence between the labels defined in the GoEmotions dataset and the terms evaluated by \citet{russell1980}. The correspondence is established not to map every single Russell term to multiple GoEmotions labels individually.}
\label{tab_word_correspondence}
\end{table*}

\section{Dependency on numbers of words}\label{sec_n_words}
\begin{figure}[h!]
\centering
\includegraphics[width=7.5cm]{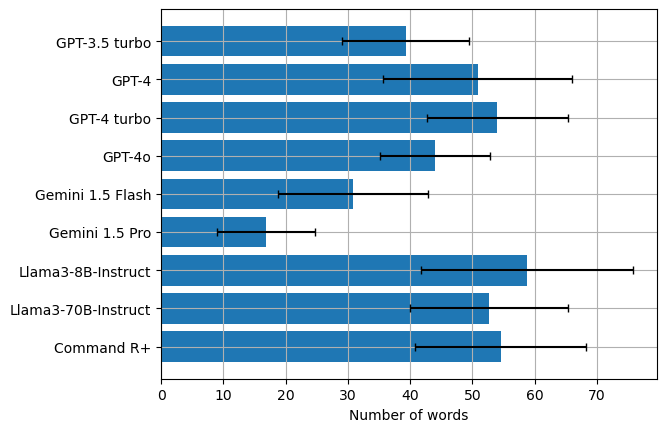}
\caption{Summary of the number of words generated by each LLM in the experiment. The bars show the mean number of words in the generated texts, and the error bars show the standard deviations.}
\label{fig_nwords}
\end{figure}

\begin{figure}[h!]
\centering
\includegraphics[width=7.5cm]{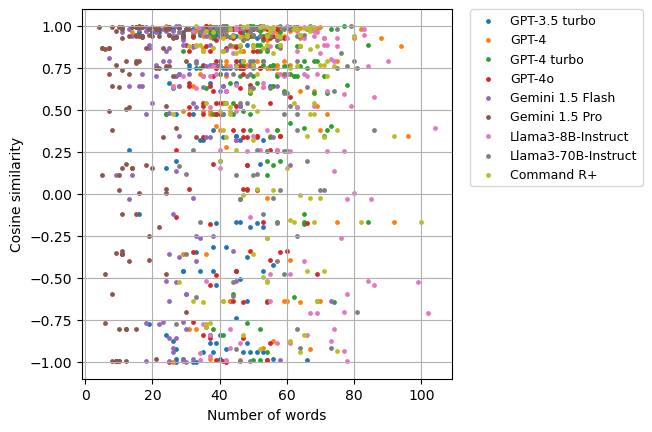}
\caption{Relation between the number of words and the cosine similarities of the generated texts. We did not observe any significant correlation.}
\label{fig_nwords_sim}
\end{figure}

We noticed that some models tend to generate more words, while others generate fewer words. The number of generated words is shown in Fig.~\ref{fig_nwords}.
We investigated whether this difference in the number of generated words affects the similarity evaluations. Figure~\ref{fig_nwords_sim} shows a scatter plot with the x-axis representing the number of words and the y-axis representing the cosine similarity.
A clear correlation, such as longer text having higher similarity, was not observed. The correlation coefficient is only 0.026, indicating no correlation. Therefore, we can conclude that there is no bias favoring some models over others in the experiment.

\section{Text Generation with Emotional States Specified by Words}\label{sec_gen_words}
\begin{figure}[h!]
\centering
\includegraphics[width=7.5cm]{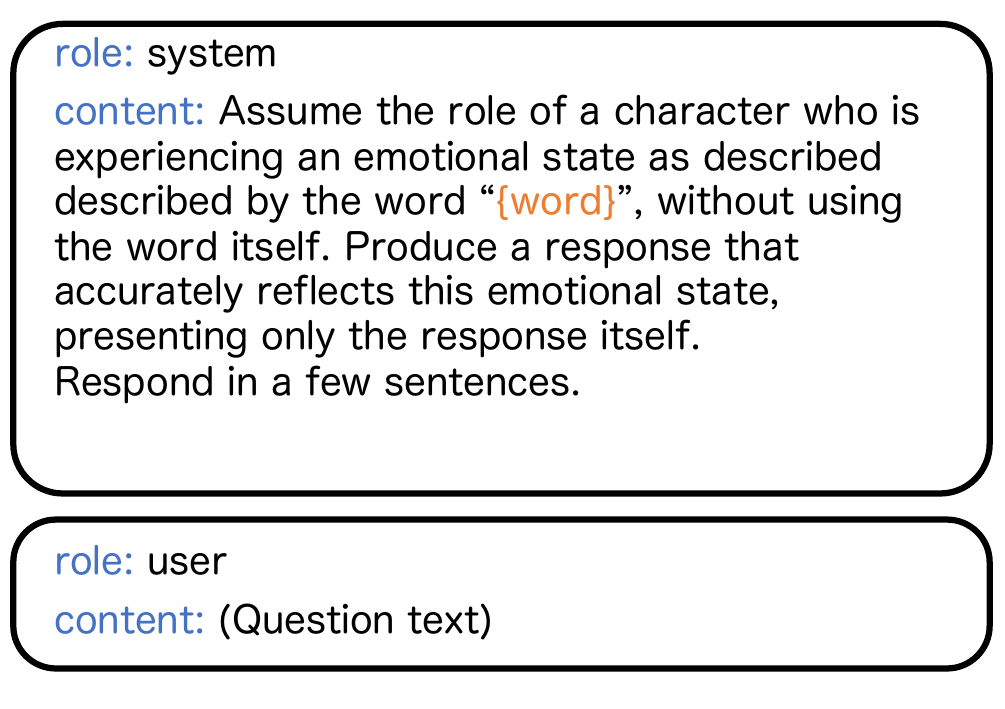}
\caption{Input prompt for text generation with a specified emotion expression described by a word.}
\label{fig_prompt_word}
\end{figure}

We conducted an experiment to generate text with emotional states described by label words from the GoEmotions dataset, using the prompt setting shown in Figure~\ref{fig_prompt_word}. To compare with the experiment presented in the main text, we selected 12 words from the 28 label words, ensuring they were as evenly distributed as possible in arousal--valence space. The selected words and their positions in the arousal--valence space are listed in Table~\ref{tab_word_pos}. 
\begin{table*}[ht!]
\small
\centering
\begin{tabular}{lrr}
\hline
\textbf{Word} & \textbf{Arousal} &\textbf{Valence} \\
\hline
pleased&0.993&$-0.119$\\
delighted&0.907&0.422\\	
astonished&0.346&0.938\\	
tense&$-0.048$&0.999\\
afraid&$-0.478$&0.878\\
frustrated&$-0.792$&0.610\\
miserable&$-0.988$&$-0.152$\\	
depressed&$-0.869$&$-0.495$\\	
bored&$-0.492$&$-0.870$\\
sleepy&0.0328&$-0.999$\\	
calm&0.722&$-0.692$\\
serene&0.854&$-0.521$\\
\hline
\end{tabular}
\caption{The list of words used in the experiment described in Appendix~\ref{sec_gen_words} to generate text with emotional states specified by the GoEmotions labels. The arousal and valence values for these words are derived from the calculations shown in Figure~\ref{fig_mapping}.}
\label{tab_word_pos}
\end{table*}

The results, showing the similarities between the specified word labels and the classified word labels in the arousal--valence space, are summarized in Table~\ref{tab_similarity_word}. 
\begin{table*}[h!]
\small
\centering
\begin{tabular}{lccccccccccc}
\hline
\textbf{Model} & \textbf{Q1}& \textbf{Q2}& \textbf{Q3}& \textbf{Q4}&\textbf{Q5}&\textbf{Q6}&\textbf{Q7}&\textbf{Q8}&\textbf{Q9}&\textbf{Q10}&\textbf{Total} \\
\hline
GPT-3.5 turbo&0.735&0.562&0.649&0.66&0.458&0.624&0.012&0.805&0.399&0.338&0.524\\
GPT-4&0.169&0.389&0.426&0.339&0.499&0.673&0.413&0.674&0.277&0.479&0.434\\
GPT-4 turbo&0.432&0.642&0.271&0.585&0.250&0.647&0.458&0.552&0.629&0.313&0.478\\
GPT-4o&0.389&0.552&0.452&0.498&0.370&0.488&0.150&0.631&0.308&0.553&0.439\\
Gemini 1.5 Flash&0.345&0.365&0.506&0.286&0.407&0.543&0.287&0.396&0.506&0.307&0.395\\
Gemini 1.5 Pro&0.268&0.379&0.244&0.072&0.278&0.477&0.311&0.273&0.104&0.295&0.270\\
Llama3-8B-Instruct&0.493&0.745&0.595&0.836&0.494&0.373&0.487&0.713&0.761&0.394&0.589\\
Llama3-70B-Instruct&0.536&0.267&0.568&0.610&0.577&0.517&0.469&0.870&0.414&0.720&0.555\\
Command R+&0.513&0.496&0.349&0.782&0.563&0.731&0.486&0.770&0.479&0.695&0.586\\
\hline
\end{tabular}
\caption{Mean cosine similarities between the emotional states specified by the word and those evaluated from the generated text for each combination of question and LLM. }
\label{tab_similarity_word}
\end{table*}

\end{document}